# Deep learning with sentence embeddings pre-trained on biomedical corpora improves the performance of finding similar sentences in electronic medical records

Qingyu Chen, Jingcheng Du, Sun Kim, W. John Wilbur and Zhiyong Lu*

National Center for Biotechnology Information, National Library of Medicine, National Institutes of Health 8600 Rockville Pike, Bethesda, MD, USA

{qingyu.chen, jingcheng.du, sun.kim, john.wilbur, zhiyong.lu}@nih.gov

*To whom correspondence should be addressed.

## Abstract

### Background
Capturing sentence semantics plays a vital role in a range of text mining applications. Despite continuous efforts on the development of related datasets and models in the general domain, both datasets and models are limited in biomedical and clinical domains. The BioCreative/OHNLP organizers have made the first attempt to annotate 1,068 sentence pairs from clinical notes and have called for a community effort to tackle the Semantic Textual Similarity (BioCreative/OHNLP STS) challenge.

### Methods
We developed models using traditional machine learning and deep learning approaches. For the post challenge, we focus on two models: the Random Forest and the Encoder Network. We applied sentence embeddings pre-trained on PubMed abstracts and MIMIC-III clinical notes and updated the Random Forest and the Encoder Network accordingly.

### Results
The official results demonstrated our best submission was the ensemble of eight models. It achieved a Person correlation coefficient of 0.8328 – the highest performance among 13 submissions from 4 teams. For the post challenge, the performance of both Random Forest and the Encoder Network was improved; in particular, the correlation of the Encoder Network was improved by ~13%. During the challenge task, no end-to-end deep learning models had better performance than machine learning models that take manually-crafted features. In contrast, with the sentence embeddings pre-trained on biomedical corpora, the Encoder Network now achieves a correlation of ~0.84, which is higher than the original best model. The ensembled model taking the improved versions of the Random Forest and Encoder Network as inputs further increased performance to 0.8528.

### Conclusions
Deep learning models with sentence embeddings pre-trained on biomedical corpora achieve the highest performance on the test set. Through error analysis, we find that end-to-end deep learning models and traditional machine learning models with manually-crafted features complement each other by finding different types of sentences. We suggest a combination of these models can better find similar sentences in practice.

## Keywords

Sentence similarity, electronic medical records, deep learning, machine learning

## Background

The ultimate goal of text mining applications is to understand the underlying semantics of natural language. Sentences, as the intermediate blocks in the word-sentence-paragraph-document hierarchy, are a key component for semantic analysis. Capturing the semantic similarity between sentences has many direct applications in biomedical and clinical domains, such as evidence sentence retrieval [1], biomedical sentence search [2] and classification [3] as well as indirect applications, such as biomedical question answering [4] and biomedical document labeling [5].

In the general domain, long-term efforts have been made to develop semantic sentence similarity datasets and associated models [6]. For instance, the SemEval Semantic Textual Similarity (SemEval STS) challenge has been organized for over five years and the dataset collectively has close to 10,000 annotated sentence pairs. In contrast, such resources are limited in biomedical and clinical domains and existing models are not sufficient for specific biomedical or clinical applications [7]. The BioCreative/OHNLP organizers have made the first attempt to annotate 1,068 sentence pairs from clinical notes and have called for a community effort to tackle the Semantic Textual Similarity (BioCreative/OHNLP STS) challenge [8].

This paper summarizes our attempts on this challenge. We developed models using machine learning and deep learning techniques. The official results show that the our model achieved the highest correlation of 0.8328 among 13 submissions from 4 teams [9]. As an extension to the models (the Random Forest and the Encoder Network) developed during the challenge [9], we further (1) applied BioSentVec, a sentence embedding trained on the entire collection of PubMed abstracts and MIMIC-III clinical notes [10], increasing the performance of the previous Encoder Network by ~13% in terms of absolute values, (2) re-designed features for the Random Forest, increasing its performance by an absolute 1.4% with only 14 features, and (3) performed error analysis in both a quantitative and qualitative manner. Importantly, the ensemble model that combines both the Random Forest and the Encoder Network further improved state-of-the-art performance -- from 0.8328 to 0.8528.

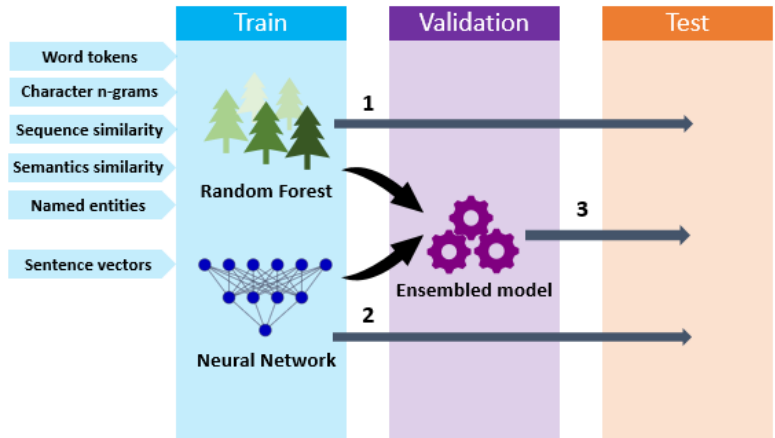

*Figure 1. An overview of our models. The Random Forest uses manually crafted features (word tokens, character n-grams, sequence similarity, semantic similarity and named entities). The feature selection of the Random Forest was done on the validation set. The Neural Network uses vectors generated by sentence embeddings as inputs. The validation set was used to monitor the early stopping process of the neural network. The ensembled (stacking) model incorporates both the Random Forest and Neural Network models. The validation set was used to train the ensembled model.*

# Methods

Figure 1 demonstrates a general overview of our models. We developed three models to capture clinical sentence similarity: the Random Forest, the Encoder Network, and the associated ensembled model. This section describes the dataset and the models in detail.

## Dataset, annotations, and evaluation metrics

The BioCreative/OHNLP MedSTS dataset consists of 1,068 sentence pairs derived from clinical notes. 750 pairs are used for the training set; 318 pairs are used for the test set. Each pair in the set was annotated by two medical experts on a scale of 0 to 5, from completely dissimilar to semantically equivalent. The specific annotation guidelines are (1) if the two sentences are completely dissimilar, it will be scored as 0; (2) if the two sentences are not equivalent but share the same topic, it will be scored as 1; (3) if the two sentences are not equivalent but share some details, it will be scored as 2; (4) if the two sentences are roughly equivalent but some important information is different, it will be scored as 3; (5) if the two sentences are mostly equivalent and only minor details differ, it will be scored as 4; and (6) if the two sentences mean the same thing, it will be scored as 5. Two medical experts annotated the dataset and their averaged scores are used as the gold standard similarity score. The agreement between the two annotators had a weighted Cohen's Kappa of 0.67 [8]. Based on the gold standard similarity score distribution, more than half of the sentence pairs have a score from 2 to 4 [8]. The Pearson correlation was used as the official evaluation metric. A good model should have high Pearson correlation with the gold standard scores, suggesting that their score distributions are consistent. More details can be found in the dataset description paper [8].

## Pre-processing sentences

We pre-processed sentences in 4 steps: (1) converting sentences into lower cases; (2) separating words joined by punctuations including "/", (e.g., "cardio/respiratory" -> "cardio / respiratory"), ".-' (e.g., "independently.-ongoing" -> "independently .- ongoing"), "." (e.g., "content.caller" -> "content . caller"), and "'" (e.g., "'spider veins'" -> "'spider veins'"); (3) tokenizing pre-processed sentences using the TreeBank tokenizer supplied in the NLTK toolkit [11]; and (4) removing punctuation and stopwords. The pre-processed sentences are used for both the Random Forest and the Encoder Network.

## The Random Forest

The Random Forest is a popular traditional machine learning model. Traditional machine learning models take human engineered features as input. They have been used for decades in diverse biomedical and clinical applications, such as finding relevant biomedical documents [12], biomedical named entity recognition [13] and biomedical database curation [14]. According to the overview of the most recent SemEval STS task [6], such traditional approaches have been still be applied by top performing systems.

In this specific task, human engineered features should be similarity measures that describe the degree of similarity between sentences, i.e., finding features that better capture the similarity between sentence pairs. Many similarity measures exist, such as the Cosine and Jaccard similarity. Zobel and Moffat [15] analyzed more than 50 similarity measures with different settings in information retrieval and found that there was no one-size-fits-all metric – no metric consistently worked better than others. We hypothesized that aggregating similarity metrics from different perspectives could better capture the similarity between sentences. We engineered features accordingly from five perspectives to capture sentence similarity: token-based, character-based, sequence-based, semantic-based and entity-based. To select the most effective features, we partitioned the official training set into training (600 pairs) and validation sets (150 pairs) and evaluated the effectiveness of the engineered features on the validation set. Ultimately 14 features achieved the highest performance on the validation set.

### Token-based features (5 features)

Token-based features consider a sentence as an unordered list of tokens; the similarity between a pair of sentences is effectively measured by the similarity between the corresponding lists of tokens. We employed 5 token-based features: (1) the Jaccard similarity [16], summarized in Equation 1, using the number of shared elements divided by the number of distinct elements in total; (2) the generalized Jaccard similarity, similar to the Jaccard similarity which considers that two tokens are the same if their similarity is above a pre-defined threshold. (We used the Jaro similarity, a string similarity measure that effectively finds similar short text [17], and empirically set the threshold to 0.6); (3) the Dice similarity [18], summarized in Equation 2; (4) the Ochiai similarity [19], summarized in Equation 3; and (5) the tf-idf similarity [20], one of the most popular metrics used in information retrieval.

$$Jaccard(X,Y) = \frac{|X \cap Y|}{|X \cup Y|}$$

*Equation 1 Jaccard similarity*

$$Dice(X,Y) = \frac{2 * |X \cap Y|}{|X| + |Y|}$$

*Equation 2 Dice similarity*

$$Ochiai(X,Y) = \frac{|X \cap Y|}{\sqrt{|X| * |Y|}}$$

*Equation 3 Ochiai similarity*

### Character-based features (2 features)

Token-based features focus on similarity at the word level. We also employ such measures at the character level. Especially, we applied the Q-grams similarity [21]. Each sentence is transformed into a list of substrings of length q (q-grams) by sliding a q character window over the sentence. We set q to 3 and 4.

### Sequence-based features (4 features)

The above measures ignore the order of tokens. We adopted sequence-based measures to address this limitation. Sequence-based measures focus on how to transform one sentence into another using three types of edits: insertions, deletions and substitutions; therefore, the similarity between two sentences is related to their number of edits. In fact, sequence-based measures are very effective in clinical and biomedical informatics; for example, they are the primary measures used in previous studies on the detection of redundancy in clinical notes [22] and duplicate records in biological databases [23]. We selected the Bag similarity [24], the Levenshtein similarity [25], the Needleman-Wunsch similarity [26] and the Smith Waterman similarity [27]. While they all measure the similarity at the sequence level, the focus varies: the Bag similarity uses pattern matching heuristics to quickly find potentially similar strings; the Levenshtein similarity is a traditional edit distance method aiming to find strings with a minimal number of edits; the Needleman-Wunsch similarity generalizes the Levenshtein similarity by performing dynamic global sequence alignment (for example, it allows assigning different costs for different operations); and the Smith-Waterman similarity focuses on dynamic sequence alignment at substrings instead of globally.

### Semantic-based features (1 feature)

The above features measure how sentences are similar in terms of syntactic structure. However, in natural language, distinct words may have close meanings; sentences containing different words or structures may represent similar semantics. For example, 'each parent would have to carry one non-working copy of the CF gene in order to be at risk for having a child with CF' and 'an individual must have a mutation in both copies of the CFTR gene (one inherited from the mother and one from the father) to be affected' contain distinct words that representing similar meanings (e.g., 'parent' vs 'the mother… and the father', and 'non-working' vs 'mutation') and the structure is rather different; however, the underlying semantics is similar. Using token-based or sequence-based features fails in this case.

We applied BioSentVec [10], the first freely available biomedical sentence encoder, on both PubMed articles and clinical notes from the MIMIC-III Clinical Database [28]. We used the cosine similarity between the sentence vectors from BioSentVec as a feature.

### Entity-based features (2 features)

In biomedical and clinical domains, sentences often contain named entities like genes, mutations and diseases [29]. The similarity of these entities embedded in sentence texts could help measure the similarity of the sentence pairs. We leveraged CLAMP (Clinical Language Annotation, Modeling, and Processing Toolkit) [30], which integrates proven state-of-the-art NLP algorithms, to extract clinical concepts (e.g. medication, treatment, problem) from the text. The extracted clinical concepts were then mapped to Unified Medical Language System (UMLS) Concept Unique Identifiers (CUI). We measured the entity similarity in the sentence pairs using Equation 4. It divides the number of shared CUIs in both sentences by the maximum number of CUIs in a sentence.

$$Entity\_Similarity(X, Y) = \frac{len(concepts_x \cap concepts_y)}{MAX(len(concepts_x), len(concepts_y))}$$

*Equation 4 Entity similarity*

In addition, we have observed that clinical notes often contain numbers other than named entities. The numbers may be expressed differently; for example, 'cream 1 apply cream 1 topically three times a day as needed' contains two numbers in different formats. We measured the similarity between numbers in two steps. First, we normalized digits to text, e.g., '24' to 'twenty-four'. Second, if both sentences contain numbers, we applied the Word Mover's Distance (WMD) [31] to measure the similarity between numbers. For other cases, i.e., if neither sentence contains numbers, the similarity is set to be 1. If only one sentence in a pair contains numbers, the similarity is set to be 0.

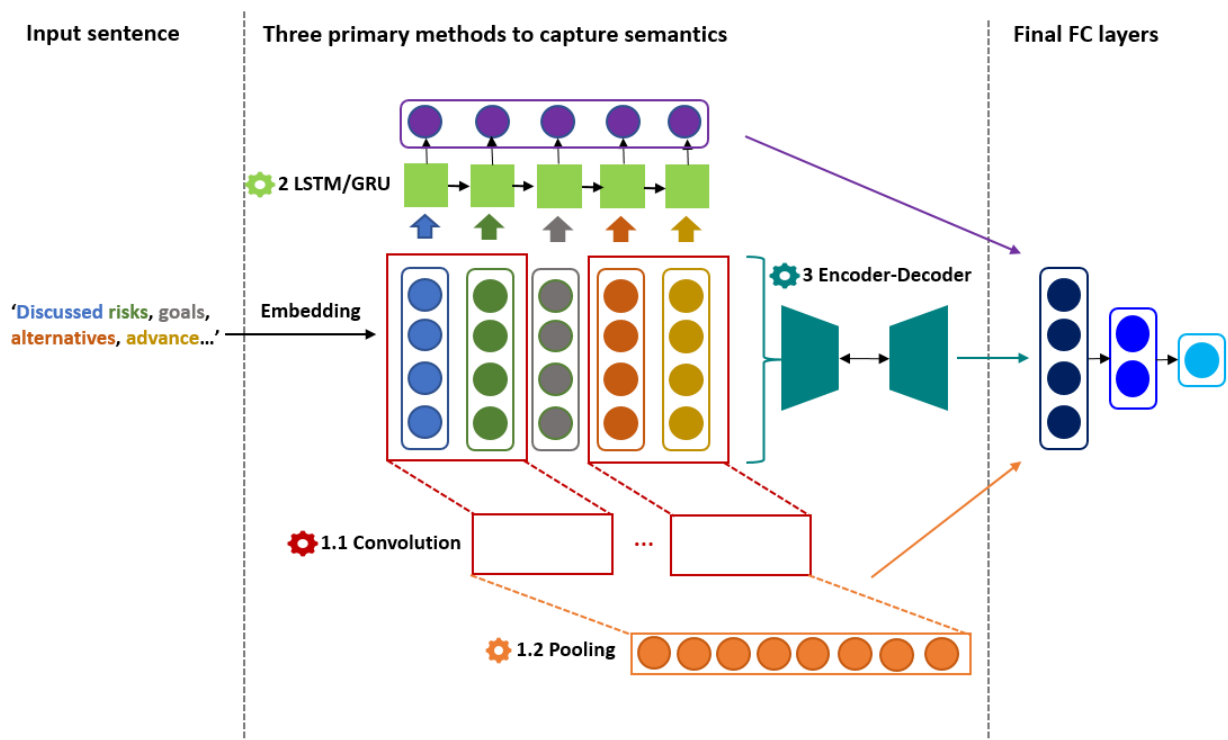

*Figure 2. Three primary deep learning models to capture sentence similarity. The first is the Convolutional Neural Network Model (1.1 and 1.2), which applies image-related convolutional processing to text. The second is LSTM or Recurrent Neural Network (2 in the figure), which aims to learn the semantics aligned with the input sequence. The third is Encoder-Decoder network (3 in the figure), where the encoder aims to compress the semantics of a sentence into a vector and decoder aims to re-generate the original sentence from the vector. FC layers: fully-connected layers. All three models use fully-connected layers as final stages.*

## Deep learning models

In contrast to traditional machine learning approaches with feature engineering, deep learning models aim to extract features and learn representations automatically, requiring minimal human effort. To date, deep learning has demonstrated state-of-the-art performance in many biomedical and clinical applications, such as medical image classification [32], mental health text mining [33], and biomedical document triage [34].

In general, there are three primary deep learning models which tackle sentence related tasks, illustrated in Figure 2. Given a sentence, it is first transformed into a 2-D matrix by mapping each word into the related word vector in a word embedding. Then to further process the matrix, existing studies have employed (1) Convolutional Neural Network (CNN) [35], the most popular architecture for computer vision, where the main components are convolutional layers (applying convolutional operations to generate feature maps from input layers) and pooling layers (reducing the number of dimensions and parameters from convolutional layers); (2) Recurrent Neural Network (RNN) [36], which aims to keep track of sequential information; and (3) Encoder-Decoder model [37], where the encoder component aims to capture the most important features of sentences (often in the format of a vector) and the decoder component aims to regenerate the sentence from the encoder output. All the three models use fully-connected layers as final processing stages.

We tried these three models and found that the encoder-decoder model (Model 3 in Figure 2) demonstrated the best performance in this task. We developed a model named as Encoder Network. It contains five layers. The first layer is the input layer. For each pair of sentences, it uses BioSentVec to generate the associated semantic vectors and concatenate absolute differences and the dot product. The next three layers are fully-connected layers each of 480, 240, and 80 hidden units. The final layer outputs the predicted similarity score. To train the model, we used the stochastic gradient descent optimizer and set the learning rate at 0.0001. The loss function is mean squared error. We also applied L2 regularization and a dropout rate at 0.5 to prevent overfitting. The training was stopped when the loss on the validation set will not decrease beyond 200 epochs. The model achieving the highest correlation on the validation set was saved.

We further developed an ensembled (stacking) model that takes the outputs of the Random Forest and Encoder Network as inputs. As explained in Figure 1, it takes the predicted scores of these two models on the validation set as inputs. Then it uses the validation set for training. We used the linear regression model that combines the scores from these two models via a linear function for ensembling.

## Results

*Table 1 Evaluation results on the official test set.*

|  | # input models | Test set correlation |
| --- | --- | --- |
| **Original submissions** |  |  |
| Random Forest (submission #1) | 1 | 0.8106 |
| Random Forest + Dense Network (submission #2) | 2 | 0.8246 |
| Ensemble model (submission #3) | 8 | **0.8328** |
| Random Forest + Encoder Network (submission #4) | 2 | 0.8258 |
| **Improved models** |  |  |
| Random Forest | 1 | 0.8246 |
| Encoder Network | 1 | 0.8384 |
| Ensemble model | 3 | **0.8528** |

Table 1 summarizes the correlation of models on the official test set. During the challenge, we made four submissions. The best model was an ensemble model using 8 models as inputs. In comparison, the performance of our models has been further improved after the challenge. The performance of the Random Forest model is improved by an absolute ~1.5%. The main difference is that we used the Cosine similarity of sentence embeddings trained on biomedical corpora as the new feature. Likewise, the performance of the Encoder Network is also improved significantly. Originally, we used Universal Sentence Encoder [38] and inferSent [39], the sentence embeddings trained on the general domain. The performance was only 0.6949 and 0.7147 respectively on the official test. Therefore, we did not make the Encoder Network as a single submission. The current Encoder Network achieved a 0.8384 correlation on the official test set, over 10% higher (in absolute difference) than the previous version. The improvement of single models further increases the performance of the stacking model. The latest stacking model is a regression model taking inputs of the single models: two using the Encoder Network with different random seeds and one using the Random Forest model. It improves the state-of-art performance by an absolute 2%.

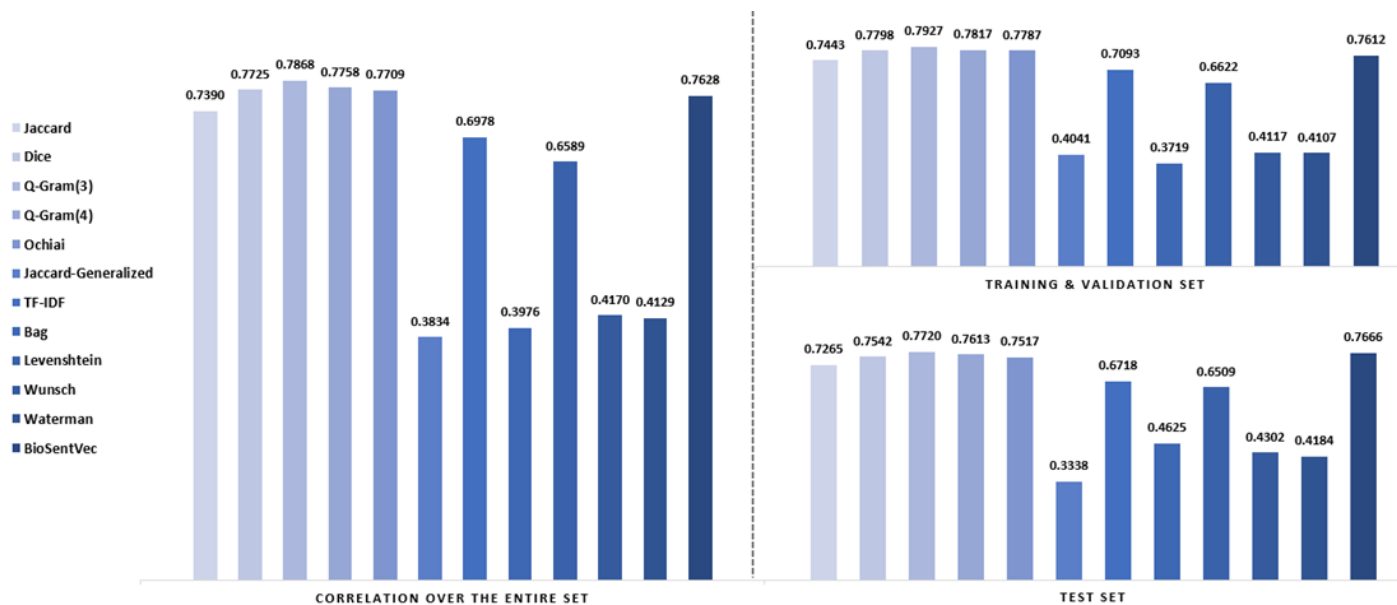

*Figure 3. Performance of an individual hand-crafted feature on the dataset. The y-axis stands for the Pearson correlation. The left shows the correlation over the entire set; the right shows the correlation over the training & validation set and the test set.*

# Discussion

## Feature importance analysis

We analyze the importance of features manually engineered for the Random Forest model. Given that most of the features can directly measure the similarity between sentence pairs, we first quantify which single feature gives the best correlation (without using supervised learning methods) on the dataset. Then we also quantify which feature has the highest importance ranked by the Random Forest.

Figure 3 demonstrates the correlation of each single feature on the entire dataset. We exclude entity-based features since they are only designed for measuring the similarity between sentences containing entities. We measure the correlation over the entire dataset and also measure the correlation over the officially released training set (train + validation) and the test set separately. The findings are two-fold. First, character-based features achieve the best correlation on all datasets: the Q-Gram (q=3) similarity had a remarkably high correlation of over 0.79 on the officially released training set and over 0.77 on the test set. This is consistent with existing studies on measuring the level of redundancy in clinical notes; for example, Zhang et al [40] found that sequence alignment incorporated with the sliding window approach can effectively find highly similar sentences. The sliding window approach essentially looks at every consecutive n characters in a sequence. This is also because the sentences are collected from the same corpus, where it is possible that clinicians use controlled vocabularies, extract sentences in templates and copy-paste from existing notes [22]. For example, we observe that snippet 'No Barriers to learning were identified' occurs frequently in the dataset.

Second, while many sentences share exact terms or snippets, the semantic-based feature, i.e., the cosine similarity generated by BioSentVec is the most robust feature. The difference between its correlation on the training set and testing set is minimal. In contrast, although Q-Gram still achieves the highest correlation in the test set, its performance drops by an absolute 2%, from 0.793 to 0.772. This shows character-based or token-based features which have lower generalization ability. The Q-Gram recognizes sentences containing highly similar snippets, but fails to distinguish sentences having similar terms but distinct semantics, e.g., 'Negative gastrointestinal review of systems, Historian denies abdominal pain, nausea, vomiting.' and 'Negative ears, nose, throat review of systems, Historian denies otalgia, sore throat, stridor.' Almost 50% of the terms in this sentence pair are identical, but the underlying semantics are significantly different. It would be more problematic when sentences are from heterogeneous sources. Therefore, it is vital to combine features focusing on different perspectives.

*Table 2 Feature ablation study on the Random Forest model. Each set of features is removed, and the difference of the performance is measured.*

|  | #features | Validation set | Test set |
|---|---|---|---|
| Full model | 14 | 0.8832 | 0.8246 |
| - Token-based | 5 | 0.8689 (-1.5%) | 0.8129 (-1.2%) |
| - Character-based | 2 | 0.8655 (-1.8%) | 0.8154 (-0.9%) |
| - Sequence-based | 4 | 0.8697 (-1.4%) | 0.8034 (-2.1%) |
| - Semantic-based | 1 | 0.8704 (-1.3%) | 0.8235 (-0.1%) |
| - Entity-based | 2 | 0.8738 (-0.9%) | 0.8150 (-0.9%) |

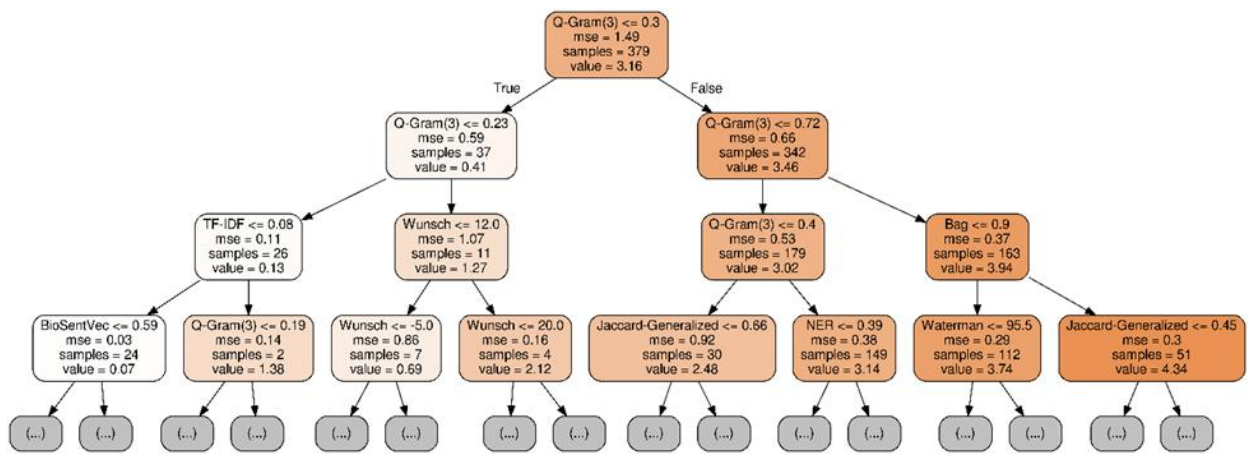

*Figure 4. A visualization of important features ranked by a tree of the Random Forest model. We randomly picked up the tree and repeated multiple times. The top-ranked features are consistent. A tree makes the decision from top to bottom: the more important the feature, the higher the ranks. In this case, Q-Gram is the most important feature. From left to right, different colors represent the sentence pairs in different degrees of similarity; darker means more similar.*

In addition, we find that sequence-based features play a vital role in the supervised setting while the correlation is low based on the previous unsupervised experiment. Table 2 shows a feature ablation study. It shows that the performance of the Random Forest model drops by an absolute 2.1% on the test set without using sequence-based features. In addition, Figure 4 visualizes the important features identified by one of the trees in the Random Forest model. We randomly selected a tree and repeated the training multiple times. The highly ranked features are consistent. Q-Gram is the most important feature identified by the model, consistent with the above results. The model further ranks sequence-based features as the second most important. For example, the Needle Wunsch similarity measure is used to split sentences into the similarity category of less than 1 or over. Likewise, the bag similarity measure is used to split sentences into the similarity category of less than ~4 or over. Therefore, while sequence-based features cannot achieve a high correlation by themselves, they play a vital role in supervised learning.

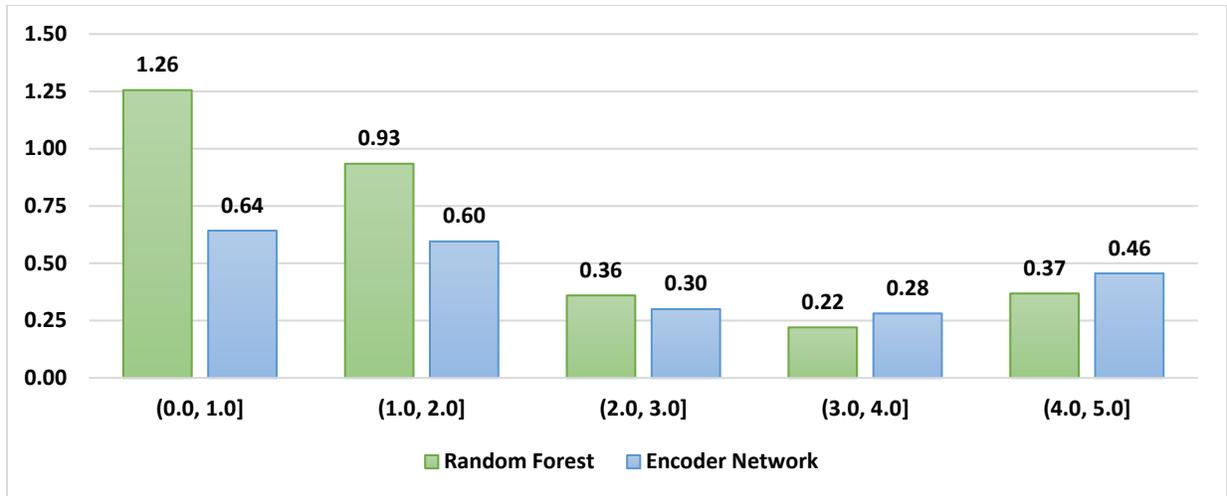

*Figure 5. The mean squared errors made by Random Forest and Encoder Model by categorizing sentence pairs into different similarity regions.*

### Error analysis

We analyze errors based on sentence similarity regions, e.g., sentences with similarities ranging from 0 to 5. For each similarity region, we measure the mean squared error produced by the Random Forest and the Encoder Network. Figure 5 shows the results. The higher value means the model made more errors. The results clearly show the Encoder Network has significantly fewer errors than the Random Forest for sentence pairs with a similarity of up to 3. This is more evident for sentence pairs of lower similarity; for example, for sentence pairs having a similarity of no more than 1, the Encoder Network had almost half of the mean squared errors. For sentence pairs of similarity over 3, the Random Forest model performed slightly better, but the difference was much smaller.

We further qualitatively demonstrated three representative cases for error analysis.

*Case 1: the Encoder Network makes more accurate predictions than the Random Forest*

- Sentence pair: 'The patient understands the information and questions answered; the patient wishes to proceed with the biopsy.' and 'The procedure, alternatives, risks, and postoperative protocol were discussed in detail with the patient.'
- Gold standard score: 2.75
- Prediction of the Random Forest model: 1.50
- Prediction of the Encoder Network: 2.12

In this case, the sentences certainly focus on different topics but they are somewhat related because they both mention there are discussions between clinicians and patients. The gold standard score is between 2 and 3. The Random Forest gave a score of 1.5, which is almost half of the gold standard score. This is because there are few shared terms in the pair; character-based or sequence-based similarities are low. The Encoder Network focuses more on the underlying semantic, thus giving a more accurate prediction.

*Case 2: the Random Forest made more accurate predictions than the Encoder Network*

- Sentence pair: 'Gradual onset of symptoms, There has been no change in the patient's symptoms over time, Symptoms are constant.' and 'Sudden onset of symptoms, Date and time of onset was 1 hour ago, There has been no change in the patient's symptoms over time, are constant.'
- Gold standard score: 3.95
- Prediction of the Random Forest model: 3.11
- Prediction of the Encoder Network: 2.43

In this case, both sentences talk about symptoms of patients. The gold standard score is 3.95. Since there are many terms or snippets shared in sentence pairs: 'onset of symptoms', 'patient's symptoms', and 'are constant'. This increases the similarity of token-based and character-based similarity measures, making the Random Forest model give a similarity score over 3. In contrast, the Encoder Network gives a score less than 2.5. For this example, the Random Forest model has a closer score than the Encoder Network. However, we believe that this example is difficult to interpret. One may argue that these sentences are both about documenting the symptoms of patients. Another may argue that the symptoms are distinct, e.g., 'gradual onset' vs 'sudden onset', and no information about onset vs 'date and time was 1 hour ago'. Besides, case 1 and 2 show that token-based and character-based features act as a double-edged sword. For pairs of high similarity, i.e. the performance of the Random Forest is improved by awarding shared terms, but it also brings false positives when sentences have low similarity but share some of the same terms.

*Case 3: the Random Forest and the Encoder Network both have significantly different scores than the gold standard*

- Sentence pair: 'Patient will verbalize and demonstrate understanding of home exercise program following this therapy session.' and 'Patient stated understanding of program and was receptive to modifying activities with activities of daily living'
- Gold standard score: 3.00
- Prediction of the Random Forest model: 1.98
- Prediction of the Encoder Network: 1.51

This is the case where both models have significantly different scores compared to the gold standard. Sentences note whether patients understand a program presumably suggested by clinicians. The gold standard score is 3, but both models give a score below 2. We believe that this is a challenging case. The pair shows some level of similarity, but in terms of the detail, they are distinct. In this specific case, the differences are not minor: the first sentence states that the therapy session will help patients understand the home exercise program, whereas the second sentence mentions that a specific patient already understands a program and is willing to change his or her activities. Case 2 and 3 in fact demonstrate the difficulty of the sentence similarity task – the relatedness is context-dependent.

## Conclusions

In this paper, we describe our efforts on the BioCreative/OHNLP STS task. The proposed approaches utilize traditional machine learning, deep learning and the ensemble between them. For post challenges, we employ sentence embeddings pre-trained on large-scale biomedical corpora and re-designed the models accordingly. The best model improves the state-of-the-art performance from 0.8328 to 0.8528.

Future work will focus more on the development of deep learning models. We plan to explore a variety of deep learning architectures and quantify their effectiveness on biomedical sentence related tasks.

# Declarations

## Ethics approval and consent to participate
N/A

## Consent for publication
N/A

## Availability of data and material
The data is publicly available via https://sites.google.com/view/ohnlp2018/home.

## Competing interests
None declared

## Funding
This research was supported by the Intramural Research Program of the NIH, National Library of Medicine, and UTHealth Innovation for Cancer Prevention Research Training Program Pre-doctoral Fellowship (Cancer Prevention and Research Institute of Texas grant # RP160015). We also thank Yifan Peng, Aili Shen and Yuan Li for various discussions. In addition, we thank Yaoyun Zhang and Hua Xu for word embedding related input.

## Authors' contributions
Conceived and designed the experiments: All
Performed the experiments: QC JD SK
Analyzed the data: QC JD SK
Wrote the paper: QC JD